\documentclass[fleqn,10pt]{pnas-new}

\templatetype{pnasresearcharticle}

\title{Differentiating Approach and Avoidance from Traditional Notions of Sentiment in Economic Contexts}

\author[1,a]{Jacob Turton}
\author[1,b,e,f]{Ali Kabiri}
\author[b,d]{David Tuckett}
\author[a,b]{Robert Elliott Smith}
\author[b,c]{David P. Vinson}

\affil[a]{Department of Computer Science, UCL}
\affil[b]{Centre for the Study of Decision-Making Uncertainty, UCL}
\affil[c]{Division of Psychology and Language Sciences, UCL}
\affil[d]{Psychoanalysis Unit, UCL}
\affil[e]{Department of Economics and International Studies, University of Buckingham}
\affil[f]{Financial Markets Group, LSE}
\affil[1]{\small \textit{Equal contribution}}

\keywords{Sentiment, Decision-making, Approach, Avoidance, Conviction Narrative Theory}

\begin{abstract}
There is growing interest in the role of sentiment in economic decision-making. However, most research on the subject has focused on positive and negative valence. Conviction Narrative Theory (CNT) places Approach and Avoidance sentiment (that which drives action) at the heart of real-world decision-making, and argues that it better captures emotion in financial markets. This research, bringing together psychology and machine learning, introduces new techniques to differentiate Approach and Avoidance from positive and negative sentiment on a fundamental level of meaning. It does this by comparing word-lists, previously constructed to capture these concepts in text data, across a large range of semantic features. The results demonstrate that Avoidance in particular is well defined as a separate type of emotion, which is evaluative/cognitive and action-orientated in nature. Refining the Avoidance word-list according to these features improves macroeconomic models, suggesting that they capture the essence of Avoidance and that it plays a crucial role in driving real-world economic decision-making.
\end{abstract}

\begin{document}

\flushbottom
\maketitle
\thispagestyle{empty}

\section*{Introduction}

The role of emotion for understanding financial markets and macroeconomic development has attracted increasing consideration in economics and finance in recent years \cite{kearney2014textual}. Mostly this attention is a consequence either of Keynes’ earlier ideas about animal spirits \cite{keynes1937general} or the more modern behavioural economics concept of the affect heuristic \cite{kahneman1982judgment, slovic2007affect}. Broadly the idea is that emotion, whether in the form of irrational mania or an unbalanced attention to loss rather than gain, tends to disrupt what would otherwise be more rational thinking so that sometimes whole markets become captured by emotional rather than rational responses to events \cite{shiller2001exuberant}. Time series measuring sentiment may therefore capture such changes but with important questions as to whether they may be cause or consequence of “real” economic events \cite{baker2006investor, baker2007investor, bordalo2018diagnostic}.

Interestingly the mainly disruptive view of the role of feelings in thinking and economic action present in economics and behavioural psychology is largely at odds with a developing line of research in neuroscience in which feeling, particularly feelings that motivate attention or action, are believed to play a crucial role in all decision-making – essentially because sub-cortical and cortical brain networks are observed to interact continuously so that all thoughts and actions are influenced by the “feeling” brain \cite{damasio2000subcortical, damasio2003feelings}. 

Conviction Narrative Theory (CNT) \cite{tuckett2017role, chong2015constructing, tuckettoecd} is a theory about decisions taken under radical uncertainty in a social context. It posits that the mental substrate underlying such decisions is a narrative — a summary representation of relevant causal, temporal, analogical, and normative information available to the decision maker - that is selected to support action although its outcome is necessarily ex ante uncertain and, therefore, potentially evokes both approach and avoidance feeling \cite{johnson2020conviction}.  

Several attempts directly to measure sentiment to assess its influence either on financial markets or broad macroeconomic indicators have been attempted. They have focused on measuring emotional words in text, defining these words using the general emotional concept of valence – broadly whether a word arouses positive or negative emotion – and have had some success e.g. \cite{chan2017sentiment, tetlock2007giving}. In this contribution we use a more nuanced method of measuring emotions in text, influenced by CNT, which posits that the sentiment changes that matter in driving an economy are those point to a shift either towards approach (increased activity) or avoidance (decreased) – Relative Sentiment Shift or RSS \cite{tuckett2014tracking}. Therefore word-lists were carefully constructed by expert judgement to represent the underlying concepts of approach and avoidance – originally conceived as excitement and anxiety, in the very specific sense as to whether the outcome was likely to be more or less conviction about action. Time series based on these word lists suggest sentiment defined in this way has significant economic effects \cite{nyman2021news,kabiri2020role}.

In this paper, a variety of statistical and machine learning techniques are used to look at what is being measured by RSS as opposed to the other more common techniques based on valence, to achieve a better understanding of what sort of emotions in text narratives seem to influence the economy. It is found not only that the RSS Avoidance word-list can be usefully enhanced and disaggregated but also that when that is done, its economic effects become particularly pronounced.  

\section{Experiment 1: Valence, Arousal and Dominance}
As mentioned, research into sentiment in financial contexts has largely focused on valence: positive and negative affect. RSS Approach and Avoidance word-lists, on the other hand, were created to capture more nuanced emotions involved in driving decision-making. To validate these claims about RSS, it is important to compare the meaning of the component words of its word-lists to alternative positive and negative sentiment word-lists and to demonstrate a difference. The Harvard General Inquirer IV (Harvard-IV) \cite{stone1966general, kelly1975computer} and Loughran-McDonald (LM) \cite{loughran2011liability, loughran2016textual} Positive and Negative sentiment word-lists have both been used to study sentiment in financial contexts, with the latter being specifically created to capture financial sentiment. They therefore make good candidates for comparison.

As a first step, the Valence, Arousal, Dominance (VAD) model of emotion \cite{mehrabian1980basic, bradley1999affective} is an obvious means of comparison as it is an established and widely accepted model of human emotion. Additionally, several large data-sets of words rated across the VAD dimensions have been produced \cite{bradley1999affective, warriner2013norms, mohammad2018obtaining} with good coverage of the RSS, LM and Harvard-IV word-lists. This allows the words of RSS, LM and Harvard-IV word-lists to be compared across these dimensions.

\subsection*{Data}
The National Research Council (NRC) VAD data-set was used as it is the largest available data-set of words rated across VAD, and there is evidence that it has good validity \cite{mohammad2018obtaining}. It contains Valence, Arousal and Dominance ratings for 20,007 words. It consists of the mean value of Valence, Arousal and Dominance for each word, averaged over a number of people’s ratings, and ranged between 0 and 1. RSS Approach and Avoidance, Harvard-IV Positive \& Negative and LM Positive \& Negative word-lists were used.

\subsection*{Method}
Valence, Arousal and Dominance values for RSS Approach \& Avoidance, LM Positive \& Negative and Harvard-IV Positive \& Negative word-lists were retrieved from the NRC data-set. A Monte-Carlo random sampling test (10, 000 re-samples) \cite{zhu2006nonparametric} was used to test significance between the RSS Approach, LM Positive and Harvard-IV Positive and RSS Avoidance, LM Negative and Harvard-IV Negative word-lists.

\subsection*{Results}
Distribution plots for RSS, LM and Harvard-IV word-lists for Valence, Arousal and Dominance can be found in the Appendix. The tables below (1\&2) display the mean values for each word-list and the significance results of RSS Approach vs. LM and Harvard-IV Positive (Table 1) and RSS Avoidance vs. Harvard-IV and LM Negative (Table 2).

\begin{table}[b]
\centering
\caption{Mean values of Valence, Arousal and Dominance for words in RSS Approach, LM Positive and Harvard-IV Positive word-lists.}
\begin{tabular}{l c c c}
\hline
& \multicolumn{3}{c}{\textbf{WORD-SET (Approach/Positive)}}\\
\textbf{FEATURE} & Harvard-IV & Loughran-McDonald & RSS\\
\hline
Valence & \textit{.757} & .852 & \textbf{.860\textsuperscript{+}}\\
Arousal & \textit{.485} & .575 & \textbf{.651\textsuperscript{*+}}\\
Dominance & \textit{.675} & \textbf{.764} & .749\\
\hline
\end{tabular}
\\
\footnotesize{\textsuperscript{*} p$<$0.01 for RSS vs LM, \textsuperscript{+} p$<$0.01 for RSS vs Harvard-IV}
\end{table}

As Table 1 shows, words in RSS Approach are significantly higher on average in Valence than those in Harvard-IV Positive ($p<0.01$) but not those in LM Positive word-lists, and significantly higher on average in Arousal than those in both Harvard-IV ($p<0.01$) and LM ($p<0.01$) Positive word-lists.

\begin{table}
\centering
\caption{Mean values of Valence, Arousal and Dominance for words in RSS Avoidance, LM Negative and Harvard-IV Negative word-lists.}
\begin{tabular}{l c c c}
\hline
& \multicolumn{3}{c}{\textbf{WORD-SET (Avoidance/Negative)}}\\
\textbf{FEATURE} & Harvard-IV & Loughran-McDonald & RSS\\
\hline
Valence & \textbf{.251} & .234 & \textit{.178\textsuperscript{*+}}\\
Arousal & .605 & \textit{.593} & \textbf{.729\textsuperscript{*+}}\\
Dominance & \textbf{.397} & .384 & \textit{.357}\\
\hline
\end{tabular}

\footnotesize{\textsuperscript{*} p$<$0.01 for RSS vs LM, \textsuperscript{+} p$<$0.01 for RSS vs Harvard-IV}
\end{table}

As Table 2 shows, words in the RSS Avoidance word-list are significantly lower in Valence on average than those in Harvard-IV (p$<$0.01) and LM (p$<$0.01) Negative word-lists. RSS Avoidance words are also significantly higher on average in Arousal than those in the Harvard-IV (p$<$0.01) and LM (p$<$0.01) Negative word-lists.

\subsection*{Discussion}
The purpose of this first experiment was to look at differences between RSS and LM \& Harvard-IV word-lists in the VAD feature space. The results found RSS Avoidance to be significantly lower in Valence and higher in Arousal than Harvard-IV and LM Negative word-lists, and RSS Approach to be significantly higher in Arousal than both but only higher in Valence than Harvard-IV. Since the RSS word-lists were originally conceived as capturing excitement and anxiety, it is not surprising that the words are particularly high in Arousal. The higher/lower Valence of RSS Approach and Avoidance respectively also suggests that RSS words tend to come from the more extreme ends of the Valence dimension.

Whilst significant differences were found in the VAD feature space, whether this is sufficient to fully differentiate RSS word-lists from their Harvard-IV and LM counterparts is doubtful. Researchers have argued that the VAD feature space is insufficient for expressing the full complexity of different types of emotions \cite{smith1985patterns}, and since RSS Approach and Avoidance are nuanced concepts, it is likely that they fall into this category. Instead, using a much larger feature space could capture the more nuanced differences between the RSS word-lists and their LM and Harvard-IV counterparts. One such space is the Binder \cite{binder2016toward} feature space which is used in the next experiment.

\section{Experiment 1b: Binder Feature Space}
The Binder feature space was developed by Binder and colleagues \cite{binder2016toward} to capture the fundamental semantic features that people use to define concepts in their minds. Through a meta-analysis they collected 65 features all demonstrated to, or strongly believed to, have neural correlates within the brain. The features range from the concrete, such as \textit{texture} and \textit{size}, to more abstract aspects such as \textit{emotions} and \textit{cognition}. The authors then collected ratings for words across the 65 features with a scale indicating how strongly the feature applied to the word (or concept it represents).

Unfortunately, the Binder data-set only exists for 535 words, severely limiting its uses for an analysis such as that carried out in Experiment 1a. However, previous research has shown that Binder features can be predicted from distributional word embeddings, allowing them to be expanded to practically any word in the English language \cite{utsumi2020exploring, turton2020extrapolating, turton2021extrapolating, chersoni2020automatic}.

The purpose of this experiment was to look for differences between the RSS, LM and Harvard-IV word-lists using the larger Binder feature space.

\subsection*{Data}
The Binder feature data-set \cite{binder2016toward} was used, consisting of 535 words scored on a scale of 0-7 across 65 features. Pre-trained Numberbatch \cite{speer2017conceptnet} embeddings were used as the input for deriving Binder features for all words in the LM, RSS and Harvard-IV word-lists.

\subsection*{Method}
To predict Binder features for all LM, RSS and Harvard-IV words, separate regression neural networks were trained (one for each of the 65 Binder features) using the human-labelled Binder data-set (see \cite{utsumi2020exploring, turton2020extrapolating} for methodological details). Following this, using the trained models, Binder features for all of the RSS, LM and Harvard-IV words were predicted. Using the predicted Binder features, Monte-Carlo randomisation tests (10,000 re-samples) were used to test the significance of differences between the RSS Approach, LM \& Harvard-IV Positive word-lists and RSS Avoidance, LM \& Harvard-IV Negative word-lists.

\subsection*{Results}
\begin{figure*}[t!]
\centering
\includegraphics[width=\linewidth]{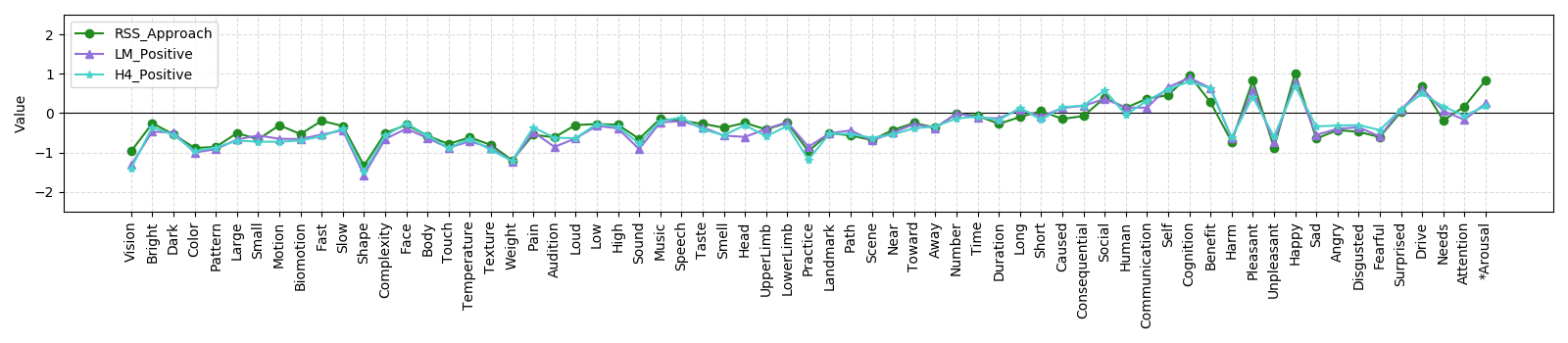}
\caption{Mean predicted feature scores for RSS Approach, LM Positive and Harvard-IV Positive word-lists, relative to mean values of Binder word-set. * p$<$0.001 RSS Approach vs both LM and Harvard-IV Positive}
\end{figure*}

\begin{figure*}[b!]
\centering
\includegraphics[width=\linewidth]{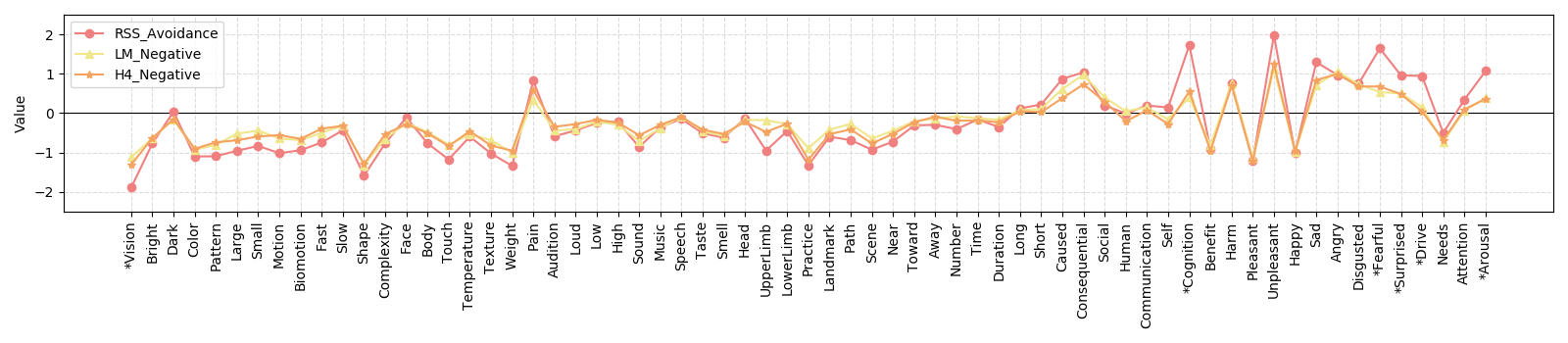}
\caption{Mean predicted feature scores for RSS Avoidance, LM Negative and Harvard-IV Negative word-lists, relative to mean values of Binder word-set. * p$<$0.001 RSS Approach vs both LM and Harvard-IV Negative}
\end{figure*}

The results (Figure 1 above) indicate that RSS Approach only differs significantly from LM and Harvard-IV Positive in the Arousal feature. Whereas RSS Avoidance (Figure 2 below) differs significantly from both LM and Harvard-IV Negative in Vision, Cognition, Fearful, Surprised, Drive and Arousal.

\subsection*{Discussion}
The Binder feature set does not have a Valence or Dominance dimension, but does have an Arousal dimension. As in Experiment 1a, both RSS Approach and Avoidance differed significantly from their LM and Harvard-IV counterparts on the Arousal Dimension. In addition, RSS Avoidance was significantly lower in the Vision feature than LM Negative and Harvard-IV Negative and significantly higher in the Cognition, Drive, Unpleasant, Fearful and Surprised features.

The lower on average values in \textit{Vision} for RSS Avoidance words may be due to them being more emotionally charged (as evidenced by lower valence) and more emotional words tend to be more abstract \cite{kousta2011representation}. The \textit{Unpleasant} feature is directly related to Valence, so with RSS Avoidance being lower in Valence it makes sense for it to be significantly higher in Unpleasantness as well. Both Fearful and Surprised features are related to Arousal, so it is not surprising that RSS Avoidance differs from its LM and Harvard-IV counterparts significantly in these features as well. More interesting are the differences in Cognition and Drive. With Cognition being "A form of mental activity or a function of the mind" \cite{binder2016toward}, it suggests that RSS Avoidance words tend to involve cognitive processes. With Drive being "someone or something that motivates you to do something" \cite{binder2016toward}, it suggests that RSS Avoidance words are action-orientated, similar to to concepts of Approach and Avoidance motivation \cite{elliot1999approach}.

Unlike RSS Avoidance, the RSS Approach word-list is poorly differentiated from its LM and Harvard-IV counterparts, even in the much larger Binder feature space, with Arousal being the only significant difference. This may be an indication that RSS Approach is a less well defined concept, that the words chosen for its word-list fail to properly capture its meaning, or that it is not very different from the existing concept of positive sentiment. Instead, it has been proposed that Approach emotions are evoked through the repelling of anxiety \cite{tuckett2017role}. This doubt repelling is unlikely to be captured by single words, but rather more complex linguistic structures or patterns.

Whilst this experiment indicates that there are significant differences in the underlying meaning of the RSS Approach and Avoidance word-lists and their LM and Harvard-IV counterparts, some of these features are themselves related to Valence \cite{warriner2013norms}. To define RSS Approach and Avoidance as different types of emotion from simple positive/negative Valence, it is important to investigate whether the other feature differences are simply due to them being from the more extreme ends of the valence dimension or due to deeper semantic differences. The next experiment tackles this by matching their valence distributions.

\section{Experiment 1c: Matched Valence}
As discussed above, whilst Experiments 1a \& 1b demonstrate that there are significant differences between RSS Approach and Avoidance and their respective LM \& Harvard-IV counterparts, it is not clear to what degree this is due to simply due to differences in Valence. If the differences between RSS and other word-lists are simply because RSS words are sampled from the more extreme ends of the Valence dimension, this damages the argument that they are more nuanced emotions. To overcome this issue, this experiment selectively sampled from the LM and Haravard-IV word-lists so as to match their Valence distributions to the RSS word-lists. These new sampled versions of LM and Harvard-IV were then compared to RSS.

\subsection*{Method}
To produce the new sampled LM and Harvard-IV lists, the Valence dimension was split into equal sized \textit{buckets} and for each bucket, the same number of words were matched between RSS Approach and LM/Harvard-IV Positive and RSS Avoidance and LM/Harvard-IV Negative word- lists. This was repeated randomly a number of times (2000) to provide estimates of the new LM and Harvard-IV word-list average feature values.

\subsection*{Results}
For the matched Valence LM \& Harvard-IV positive word-lists, RSS Approach remained significantly higher in Arousal (p$<$0.001), remaining the only feature significantly different between the word-lists.

For matched Valence LM \& Harvard-IV negative word-lists RSS Avoidance was no longer significantly different in Unpleasant or Vision, but remained significantly different (p$<$0.001) in Cognition, Drive, Fearful, Surprised and Arousal (see Figure 3 below).

\begin{figure*}[b!]
\centering
\includegraphics[width=\linewidth]{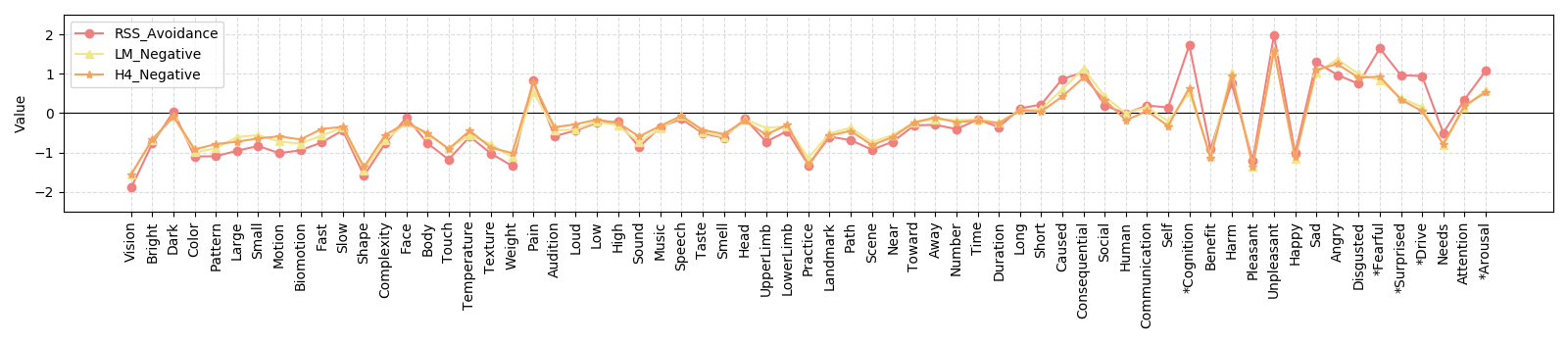}
\caption{Mean predicted feature scores for RSS Avoidance and matched Valence LM Negative and Harvard-IV Negative word-lists, relative to mean values of Binder word-set. * p$<$0.001 RSS Approach vs both LM and Harvard-IV Negative}
\end{figure*}

\subsection*{Discussion}
As the results show, by matching for Valence, RSS Approach remains significantly higher in Arousal than its LM and Harvard-IV counterparts. For RSS Avoidance, the Vision and Unpleasant features are no longer significantly different, but the Cognition, Drive, Fearful, Surprised and Arousal features remain so. This suggests that these differences are deeper than simply a result of sampling words from the more extreme ends of the Valence dimension. Together with Experiment 1b, these results also demonstrate that RSS Avoidance has more differences compared to its LM and Harvard- IV counterparts than RSS Approach (which is only different across a single feature). This suggests RSS Avoidance may be better defined as a separate type of sentiment. As such, RSS Avoidance becomes the focus of the final parts of this paper.

It is worth now taking a closer look at the definition of the features indicated as significantly different for RSS Avoidance to better understand the differences:

In the Binder paper, Cognition is defined as relating to abstract concepts that involve cognitive processes such as thinking and planning. They indicate that it includes both working-memory type processes, which are important for reasoning and decision-making \cite{diamond2013executive} and cognitive-control/executive processes which involve the cognitive, purposeful control of behaviour \cite{diamond2013executive}.

Binder and colleagues define Drive in terms of Maslow’s hierarchy of needs \cite{maslow1943theory}. The most basic of these needs must be met for survival, whereas others are desirable to meet homeostasis (biological balance). It has been proposed that meeting these needs is the basis of human behavioural motivation and therefore a driver of behaviour. This has strong parallels to the concept of Approach and Avoidance motivation proposed by Elliott \cite{elliot1999approach} and others, which also has been framed in terms of motivating behaviour to meet fundamental needs.

Both Fearful and Surprised are defined by Binder and colleagues in terms of Ekman’s six basic emotions \cite{ekman1971constants, ekman1997universal}. Fear is a reaction to a stimulus that is perceived as dangerous, and motivates the protective “fight or flight” response \cite{fox2008emotion}. Surprise is the reaction to a stimulus that often results in the startle response \cite{fox2008emotion}. Unlike Fear, Surprise can be either positively or negatively valenced, but extremely negative Surprise may be considered Fear \cite{noordewier2013valence}.

\section{Experiment 2: A Closer Look at RSS Avoidance}
Experiments 1a, b \& c demonstrated that there are significant differences between RSS Avoidance and its Harvard-IV and Loughran-McDonald counterparts across a number of semantic features, even when matched for Valence. However, this raises the question of how homogeneous the RSS Avoidance words are across these features, or whether there are different sub-groups of words within the word-list and whether they have different semantic properties. This experiment approaches this issue by using unsupervised statistical methods to identify any patterns across the features that define RSS Avoidance (Cognition, Drive, Fearful, Surprised and Arousal).

\subsection*{Method}
Using the five features found to be important for RSS Avoidance: Fearful, Surprised, Cognition, Drive and Arousal, correlations between the features across the words were calculated as an initial exploratory step. Following this, a Principal Component Analysis (PCA) was carried out to see how the features loaded onto a smaller number of dimensions. Finally, an unsupervised clustering method (k-means clustering) was used to cluster the words. An unsupervised method was used to keep any author preconceptions and biases out of the process of splitting the words.

\subsection*{Results}
The initial exploratory correlation analysis indicated two “groups” of features with Surprised and Fearful being correlated, Cognition and Drive being correlated and Arousal somewhat straddling the two groups. (Figure 4 below).

\begin{figure}[ht]
\centering
\includegraphics[width=.5\linewidth]{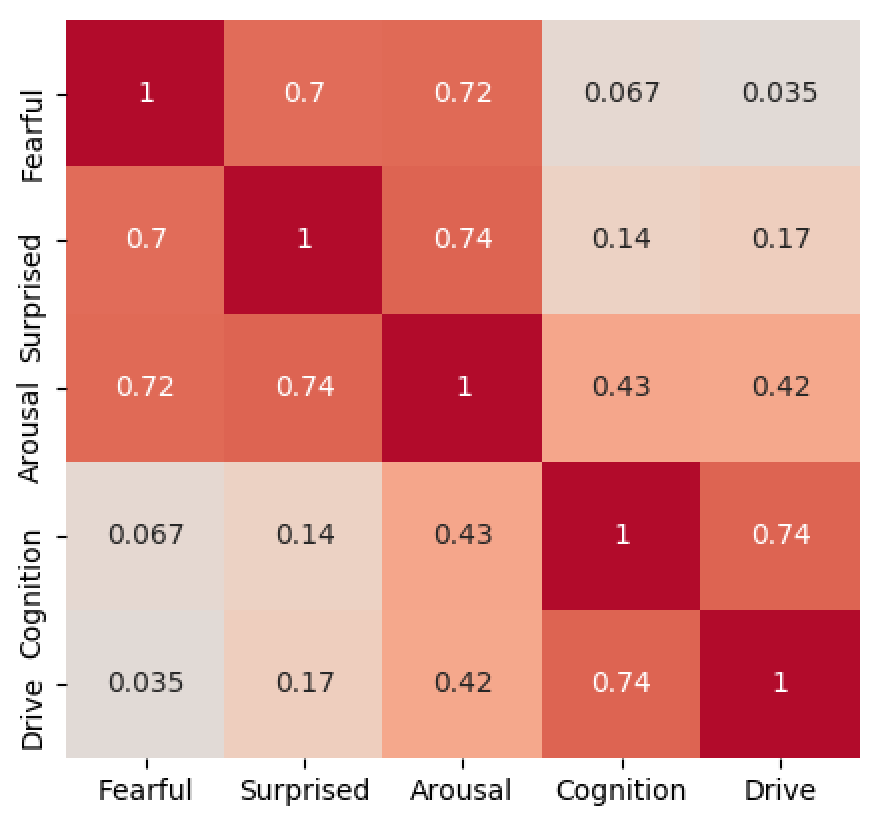}
\caption{Pearson correlations between features for RSS Avoidance words}
\label{fig:corr}
\end{figure}

The correlations indicated that a PCA would be appropriate to reduce the feature space to two dimensions. Since Arousal was at least moderately correlated with both groups it was left out. Therefore, using Fearful, Surprised, Cognition and Drive, a two-component PCA captured 88\% of variance across the features within the word-list, with one component loading higher for Fearful and Surprised and the other for Cognition and Drive. Figure 5 below shows the RSS Avoidance words plotted in the resultant 2-dimension PCA space, with the results of the unsupervised k-means clustering included.

\begin{figure}[ht]
\centering
\includegraphics[width=.6\linewidth]{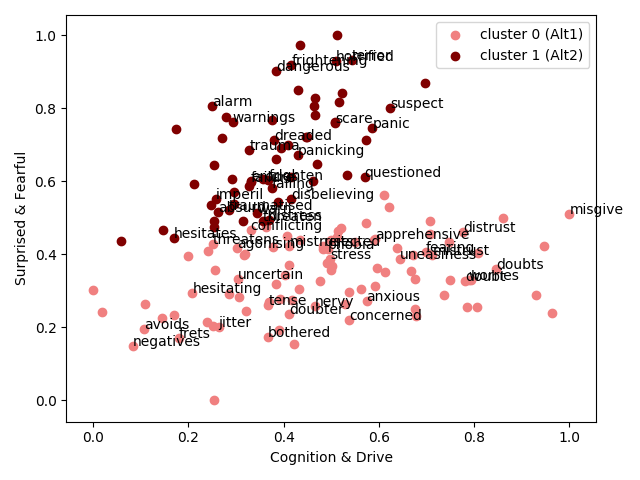}
\caption{RSS Avoidance words across 2 PCA dimensions and clustered}
\label{fig:clus}
\end{figure}

\subsection*{Discussion}
The purpose of this experiment was to investigate how semantically homogeneous (across the features of interest) the RSS Avoidance word list is. The results demonstrated that two clear lower dimensions exist within the word-list, one loading on Fearful and Surprised and the other loading on Cognition and Drive. Clustering the words in this space gave two distinct groups, with each one having to have words higher in each of the dimensions. Given the definitions of these features earlier in the paper, these groupings make sense. One appears to contain words related to a negative reaction to a stimulus; Fear and Surprise, whereas the other contains words related to higher-order mental processes controlling behaviour to meet needs; Cognition and Drive. These are quite distinct types of negative emotion.

However, simply identifying these two groups and speculating on their differences is of limited use. Whilst it is interesting to see that two different types of words appear to be present in the RSS Avoidance word-list, any implications of this are unanswered. In particular, the questions arise: are both types of words important for RSS? Does one group perform better than the other for explaining economic outcomes? And even: does getting rid of one group improve RSS overall? The final experiment explores this question by comparing the performance of these two new Avoidance word-lists to the original RSS Avoidance when used as inputs in econometric models.

\section{Experiment 3}
The purpose of this third experiment was to compare the performance of the two different groups of RSS Avoidance words with each-other, the original full RSS Avoidance word-list and the LM word-lists for their effect on economic outcomes. Using the various word-lists, indices of sentiment were derived from the Reuters historical news archive - a digitised news data-set - and used as inputs for an econometric model. The performance of the different word-lists could then be compared. 

\subsection*{Data}
The Reuters digital news archive was used, covering all Reuters news articles tagged New York and Washington D.C. from Dec 1995 to Aug 2016. LM Positive \& Negative, Original RSS Approach \& Avoidance and Alt1 \& Alt2 Avoidance word-lists were used.

\subsection*{Method}
For each article, where (\(p\)) is the number of positive/approach words, (\(n\)) is the number of negative/avoidance words and (\(t\)) is the total number of words in the article, sentiment (\(s\)) was calculated as (1) below.

\begin{equation}
s = \frac{p - n}{t}
\end{equation}

Per article sentiment was aggregated by the month resulting in a sentiment time series for each of the sentiment indices being tested: Original RSS, RSS Alt1, RSS Alt2 and LM.

In order to determine the impact of measures of sentiment on the real economy, a vector error correction model was estimated that contained the following variables at a monthly frequency: the (natural) logarithm of real GDP (RGDP), the logarithm of Real investment (RINV) the logarithm of the Standard and Poor’s 500 stock market index (SP), the Fed funds rate (FEDFUND), the PCE inflation rate, the University of Michigan inflation expectations survey rate (INFEXP), the TED spread (TED), The VIX index (VIXCLS), the GZ credit spread (GZSPR), the excess bond premium (EBPOA) and the four measures of sentiment (Original RSS, Alt1, Alt2, LM). The credit risk spread and the excess bond premium are those computed in \cite{gilchrist2012credit}.

Of the eleven time series in the model, four were found to be stationary: inflation expectations, VIX, the GZ spread, and the excess bond premium. For all other time series, the augmented Dickey-Fuller unit root test could not be rejected. A mixed vector error correction model was estimated. Cointegration among the seven time series that contain a unit root was tested for using the Johansen method\cite{Johansen1995integ}, and it was found that there were two cointegrating relationships. A vector error correction model was then estimated with six cointegrating relationships. The two cointegrating relationships that were defined for the seven non-stationary variables excluded the four stationary variables. The other four cointegrating relationships were identities for each of the stationary time series. A lag length of 2 was chosen by minimising the Bayesian Schwarz information criteria. The model was estimated and results were used to identify orthogonalized shocks using a Cholesky decomposition of the variance-covariance matrix.

\begin{figure*}[b!]
\centering
\includegraphics[width=\linewidth]{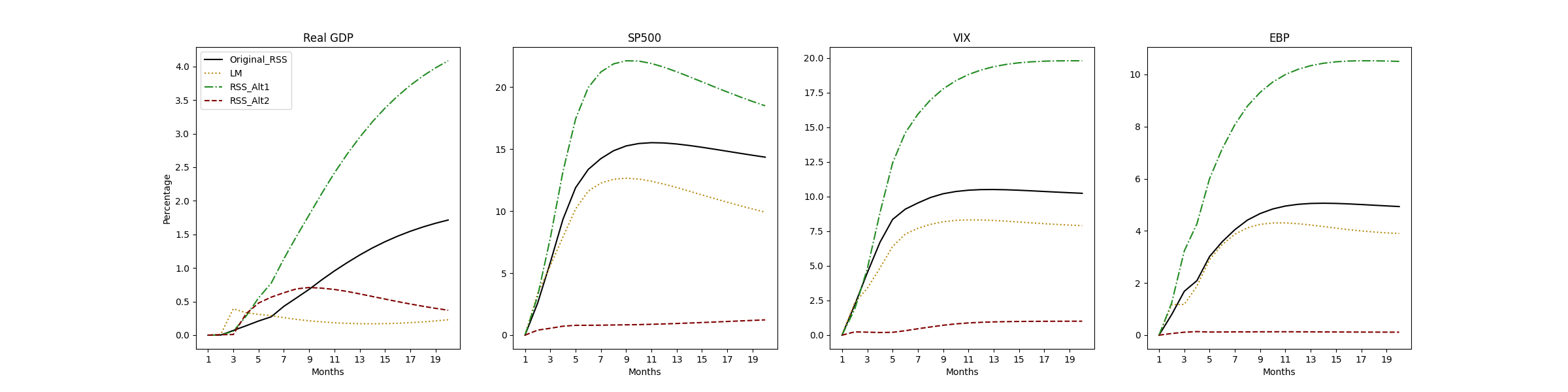}
\caption{Contribution of shocks to forecast error variance for Real GDP, S\&P500, VIX and Excess Bond Premium}
\end{figure*}

The shock to sentiment was identified using an orthogonalized decomposition of the variance-covariance matrix obtained by estimating a vector error correction model (VECM). The sentiment series was ordered last in the model. This had a number of implications to the interpretation of the shocks that were identified. First, the sentiment shock was orthogonal to all other shocks in the model. That is, the sentiment shock was orthogonal to the shock to real GDP, shock to real investment, shock to the stock market, shock to the short-term interest rate, shock to inflation, shock to inflation expectations, shocks to the TED Spread, shocks to VIX, shocks to credit risk spreads (GZ),and shocks to the EBP. There is no structural interpretation to these shocks, but it is reasonable to expect that these shocks include aggregate demand and aggregate supply shocks, along with credit market and monetary shocks.

As the sentiment variable was ordered last, the identified sentiment shock does not have an immediate impact on any of the variables in the econometric model. It is possible however that the sentiment that is measured is contaminated by “news”. By ordering sentiment last we are allowing the “news” contamination in measured sentiment to be filtered away leaving pure sentiment. The reasoning is twofold: first, news about the economy is internalized in stock and credit markets. The business-sentiment shock that is identified is orthogonal to shocks to stock and credit markets. Second, the sentiment shock only affects the other variables in the system with a delay. Thus we are identifying shocks to the slow moving (or long-run component) of business-sentiment and not the short-run component. Our identification is that the long-run component of sentiment is not contaminated by news. Given the ordering, the sentiment shock is interpreted as a “pure” sentiment shock after controlling for shocks to all other variables.

The forecast error variance decompositions in Fig 5 for the key variables shows that the identified sentiment shock accounts for up to four percent of the forecast error variance for Real GDP, for up to twenty-two percent of the forecast error variance for the S\&P 500 stock index, for up to twenty percent of the forecast error variance of VIX, and up to ten percent of the forecast error variance for the EBP, all over the medium to long term (20 months).

Figure 6 reports the impulse responses of Real GDP, S\&P500 stock index, VIX and EBP to a one SD shock from RSSALT1 identified from the vector error correction model. The confidence intervals are constructed using the method of Hall with 1000 bootstrap replications and the bands represent a 95\% confidence interval \cite{Hall1992} The full process was repeated for the other three indexes and are reported in the next section.

\subsection*{Results}
The results show that shocks to RSS Alt1 have clear impacts on several key economic and
financial variables over the medium to long run. The magnitude of the responses in Fig 6 show that impacts on financial risk variables reach their peak around 5 months and the scale of these are relatively large. The effect on the SP 500 is 2.7\% at 12 months. For Real GDP the effect is 0.15\% at 15 months (see Fig 6). What is also of interest is the way the new index based on cognition and drive has larger impacts than its counterpart based on Fear and Surprised and, the LM and RSS indexes. Figs 8,9,10 and 11 in the appendix show that the scale of the impacts are much less for LM, reduced for Original RSS and weak for the RSS Alt2 index.
For example, the impacts on the SP500 are forty-two per cent larger and, for the VIX and the excess bond premium (EBP) fifty per cent larger, than the LM index.

\begin{figure}[ht]
\centering
\includegraphics[width=.8\linewidth]{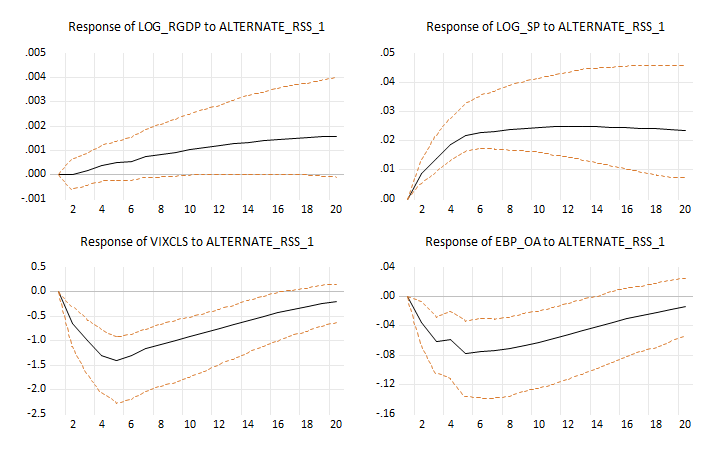}
\caption{Orthogonalised Impulse responses of log Real GDP, SP500, VIX and EBP from a one SD shock to AltRSS1}
\label{fig:oirf}
\end{figure}

\subsection*{Discussion}
The results highlight three major observations. The first, that our separation of the word lists create
two alternate lists which, when converted to sentiment indexes have markedly different effects on the economy. This confirms that they identify different information in news text data, and the group related to ‘Cognition and Drive’ (Alt1) has the most pronounced effects. The other group- based on ‘Fear and Surprised’ is significantly weaker.

Second, the overall impacts of our new index on the economy are statistically significant, and are economically meaningful. Third, the new word-list for Cognition and Drive improves upon LM and original RSS. Given the widespread use of the LM index 13 the method reveals a useful method to better identify the role of human emotion in financial and economic news text data as related to economic activity.

\section{General Discussion}
This research began as an exploratory study into the underlying meaning of words chosen to represent CNT's Approach and Avoidance concepts when compared to existing word-lists based on positive and negative sentiment. It was found that RSS Avoidance differed significantly across the features of Fearful, Surprised, Cognition, Drive and Arousal, even when controlling for Valence. RSS Approach, on the other hand, was less well defined as a separate type of emotion, only significantly differing to lists of negative sentiment on Arousal.

A closer look at the CNT Avoidance list indicated two different types of words were present; those higher in Cognition \& Drive and those higher in Surprise \& Fearful. Using these groups as separate word-lists, econometric modelling indicated that the words higher in Cognition and Drive were superior at explaining economic outcomes than both original RSS and the words higher in Surprise and Fearful, as well as LM. This indicates that if CNT Avoidance does indeed capture the sentiment of decision-making, it is likely that it involves cognitive, action-orientated processes rather than either reactive fear/surprise or generic negative sentiment (as represented by LM).

\section*{Acknowledgments}

This research was supported by grants to David Tuckett and the UCL Centre for the Study of Decision-Making Uncertainty from the Institute of New Economic Thinking (IN011-00025, IN13-00051 and INO16-00011), the Eric Simenhauer Foundation of the Institute of Psychoanalysis (London), UKRI (EPSRC grant reference EP/P016847/1) and the ESRC-NIESR Rebuilding Macroeconomics network).

\bibliography{main}

\clearpage
\section*{APPENDIX}

\begin{figure*}[ht]
\centering
\includegraphics[width=\linewidth]{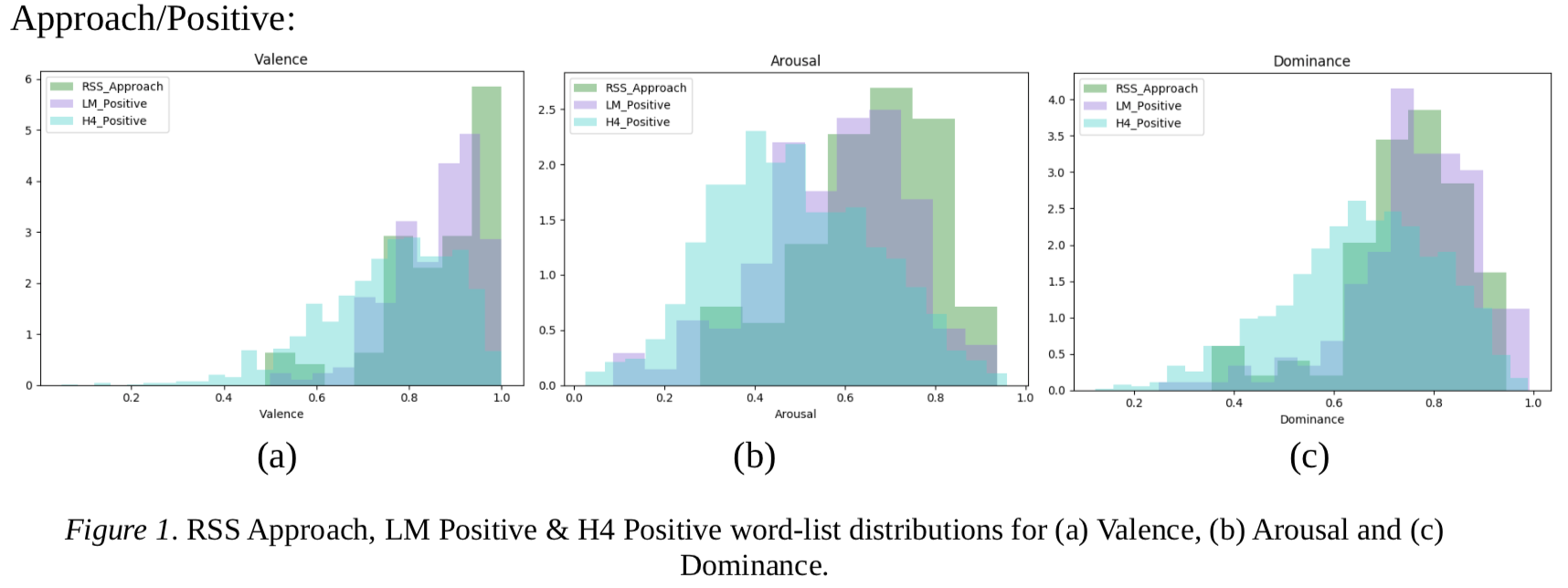}
\caption{}
\end{figure*}

\begin{figure*}[ht]
\centering
\includegraphics[width=\linewidth]{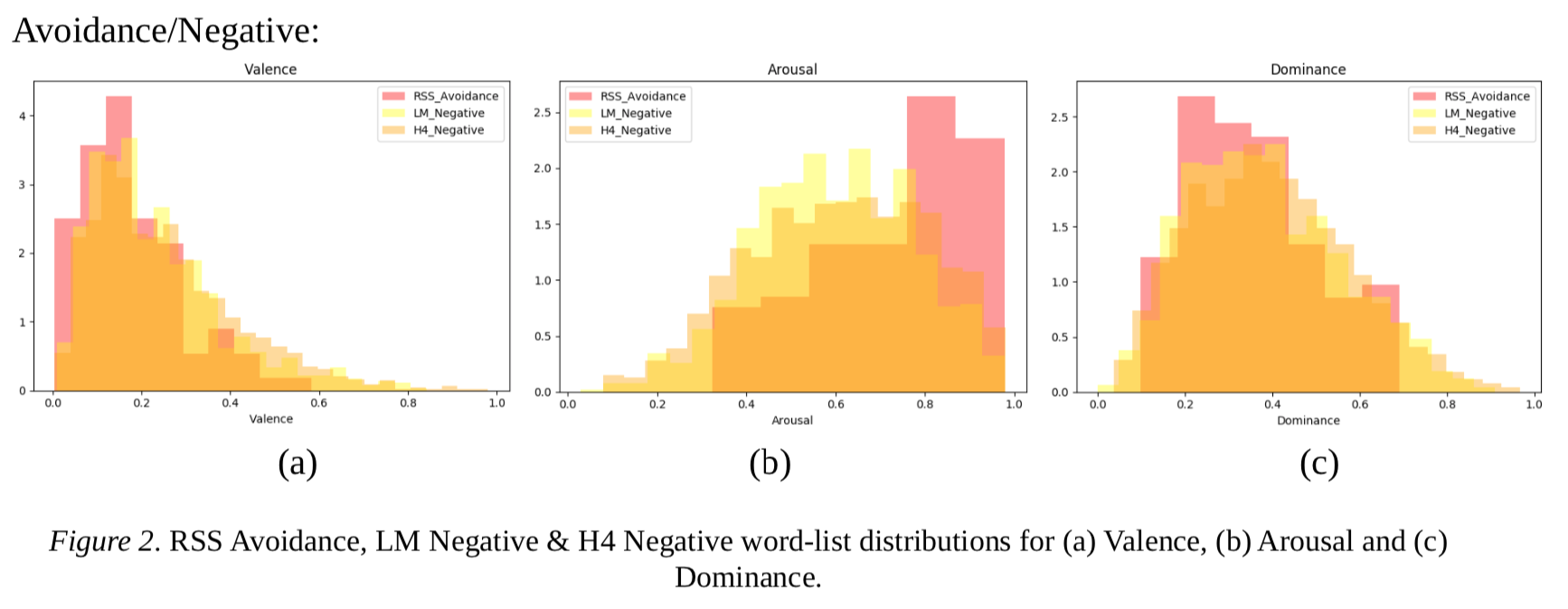}
\caption{}
\end{figure*}

\begin{figure*}[!b]
\centering
\includegraphics[width=\linewidth]{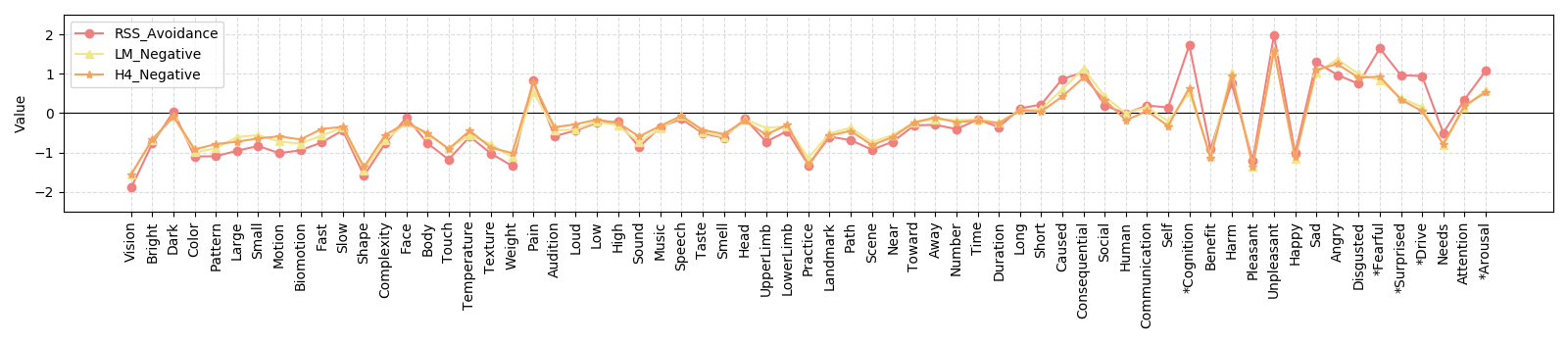}
\caption{Mean predicted feature scores for RSS Avoidance and matched Valence LM Negative and Harvard-IV Negative word-lists, relative to mean values of Binder word-set. * p<0.001 RSS Approach vs both LM and Harvard-IV Negative}
\end{figure*}

\begin{figure*}[b!]
\centering
\includegraphics[width=\linewidth]{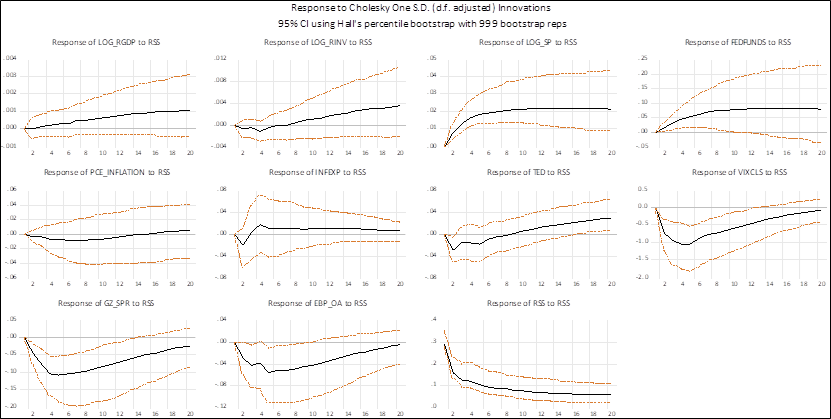}
\caption{Impulse responses for a one SD shock to RSS sentiment}
\end{figure*}

\begin{figure*}[t!]
\centering
\includegraphics[width=\linewidth]{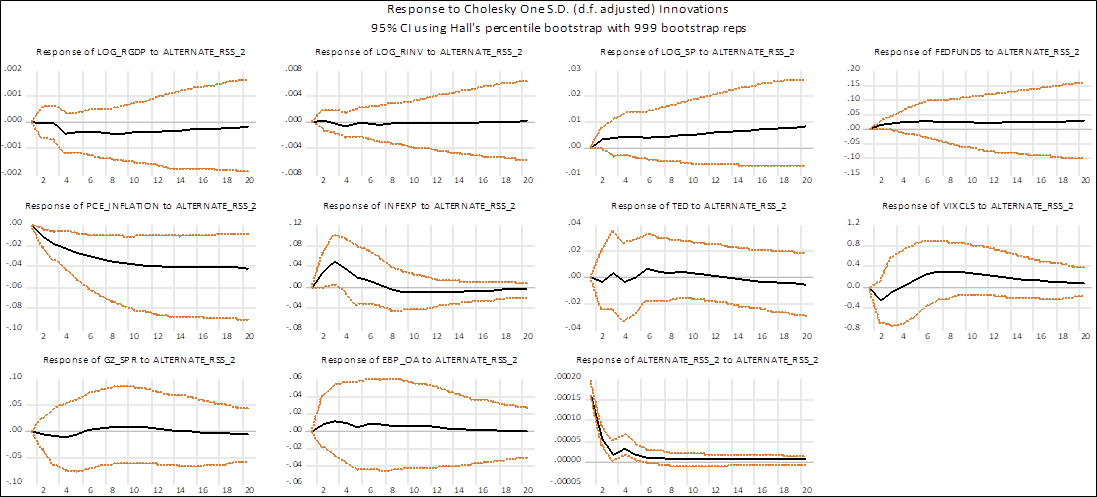}
\caption{Impulse responses for a one SD shock to Alternative RSS 2 sentiment}
\end{figure*}

\begin{figure*}[b!]
\centering
\includegraphics[width=\linewidth]{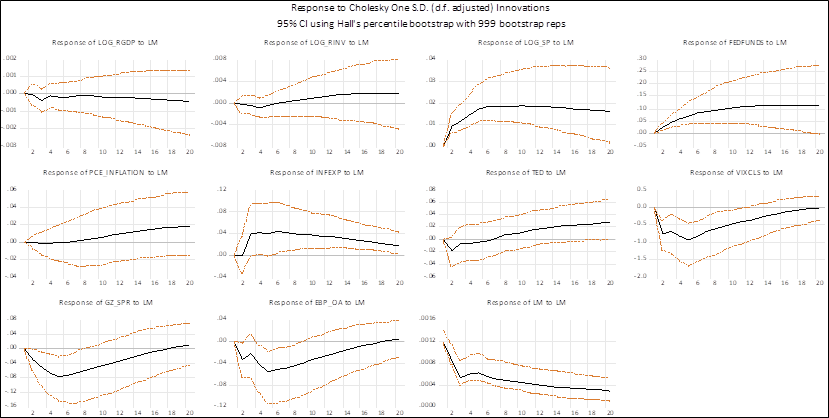}
\caption{Impulse responses for a one SD shock to LM sentiment}
\end{figure*}

\begin{figure*}[t!]
\centering
\includegraphics[width=\linewidth]{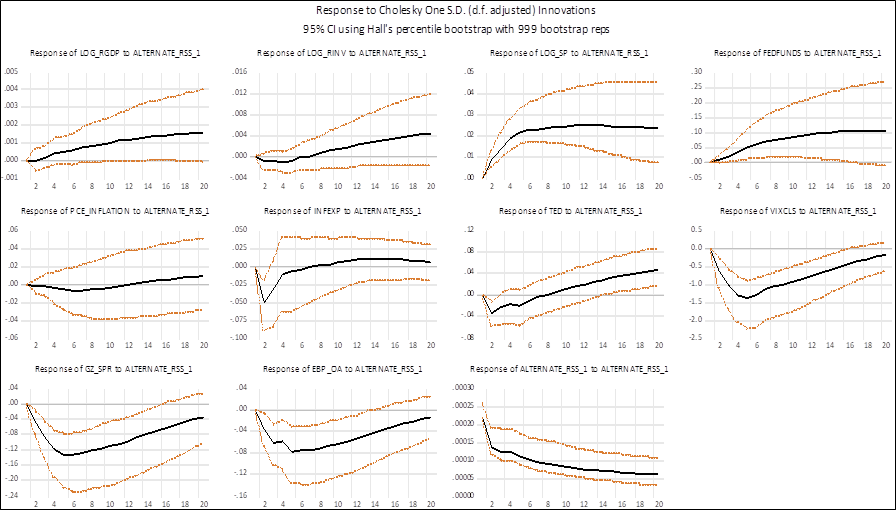}
\caption{Impulse responses for a one SD shock to Alternative RSS 1 sentiment}
\end{figure*}

\end{document}